\documentclass{article}

\usepackage{arxiv}

\usepackage[utf8]{inputenc}
\usepackage[T1]{fontenc}
\usepackage{hyperref}
\usepackage{url}
\usepackage{booktabs}
\usepackage{amsfonts}
\usepackage{amsmath}
\usepackage{amssymb}
\usepackage{nicefrac}
\usepackage{microtype}
\usepackage{graphicx}
\usepackage{xcolor}
\usepackage{capt-of}
\usepackage{natbib}
\usepackage{doi}
\usepackage{enumitem}
\usepackage[nameinlink,noabbrev]{cleveref}

\title{Anything2Skill: Compiling External Knowledge into Reusable Skills for Agents}

\author{
  Qianjun Pan$^{1}$, Yutao Yang$^{1}$, Junsong Li$^{1}$, Jie Zhou$^{1,2}$\footnotemark[1], Kai Chen$^{2}$, Xin Li$^{2}$, Qin Chen$^1$, Liang He$^1$ \\
  $^1$ School of Computer Science and Technology, East China Normal University, Shanghai\\ 
  $^2$ Shanghai AI Laboratory \\
  \texttt{\{jzhou, qchen, lhe\}@cs.ecnu.edu.cn} \\
  \textcolor{red}{\url{https://github.com/ECNU-ICALK/AutoSkill}} 
}

\renewcommand{\shorttitle}{Anything2Skill}

\begin{document}
\maketitle

\begingroup
\renewcommand{\thefootnote}{\fnsymbol{footnote}}
\footnotetext[1]{Corresponding Author}
\endgroup

\begin{abstract} 
Retrieval-augmented generation (RAG) enables agents to access external knowledge at inference time, but it primarily retrieves fragmented declarative evidence, leaving agents to repeatedly infer task procedures from passages, manuals, examples, logs, or trajectories. This raises a fundamental question: \textbf{can skills extracted from external knowledge bases be installed into an agent, enabling it to rapidly approximate domain expertise?} In this paper, we propose \textsc{Anything2Skill}, a taxonomy-guided framework that compiles heterogeneous external knowledge into reusable, retrievable, and executable skills for agents. Given a corpus of knowledge records, \textsc{Anything2Skill} first decomposes each record into evidence windows and performs plan-and-expand skill extraction under a skill-tree prior. The extracted candidates are then converted into structured skill contracts that specify invocation conditions, contraindications, action moves, workflow steps, constraints, output specifications, supporting evidence, and confidence scores. To construct a deployable procedural memory, \textsc{Anything2Skill} manages the extracted skills in a persistent SkillBank through taxonomy-aware compilation, registry-level reconciliation, lifecycle tracking, versioned updates, and visible skill-tree projection. At inference time, agents retrieve both task-specific passages from the original knowledge base and relevant procedural skills from the SkillBank, allowing RAG to provide declarative evidence while compiled skills provide reusable procedural guidance. Experiments on qsv and GitHub-CLI show that \textsc{Anything2Skill} combined with RAG achieves 98.85\% and 94.10\% success rates, respectively, substantially outperforming RAG-only agents. These results suggest that compiling latent procedural knowledge into explicit skills is an effective way to extend retrieval-augmented agents from knowledge access toward capability reuse.
\end{abstract}



\section{Introduction}
\label{sec:introduction}

Large language models (LLMs) are increasingly deployed as autonomous agents that reason over external information, invoke tools, interact with environments, and complete multi-step tasks. A central challenge for such agents is how to acquire and reuse knowledge beyond the parametric memory of the underlying model. Retrieval-augmented generation (RAG) has emerged as a widely adopted solution by enabling models to access external, updateable, and interpretable knowledge at inference time \citep{lewis2020rag,guu2020realm,karpukhin2020dpr}. By retrieving relevant documents, passages, web pages, or API descriptions, RAG substantially improves factual grounding and supports knowledge-intensive reasoning across open-domain question answering, long-form generation, and tool-use scenarios \citep{izacard2021fid,borgeaud2022retro,izacard2023atlas,asai2024selfrag}.

Despite its effectiveness, existing RAG systems primarily treat external knowledge as declarative evidence. Retrieved passages tell the agent what information is relevant, but they usually do not directly specify how a task should be performed. As a result, agents must repeatedly reconstruct procedural knowledge from fragmented evidence, such as manuals, examples, dialogue histories, execution logs, or task trajectories. This limitation is especially salient in agentic settings, where success often depends not only on retrieving the right facts, but also on following the right procedures: when to invoke a tool, which action sequence to execute, what constraints to satisfy, what intermediate artifacts to produce, and which failure cases to avoid. Even advanced retrieval-augmented reasoning methods that interleave retrieval with generation or reflection \citep{trivedi2023ircot,jiang2023flare,khattab2022dsp,asai2024selfrag} still require the model to infer such procedures at task time, rather than storing them as reusable capabilities.

In parallel, recent work on LLM agents has shown that explicit skills, tools, memories, and executable routines are crucial for long-horizon problem solving \cite{yang2026autoskill}. ReAct interleaves reasoning and acting to improve interactive decision making \citep{yao2023react}; Toolformer, Gorilla, ToolLLM, and related systems improve the ability of LLMs to call external APIs and tools \citep{schick2023toolformer,patil2024gorilla,qin2024toolllm}; Reflexion and Generative Agents demonstrate the value of externalized memory and self-reflection for agent improvement \citep{shinn2023reflexion,park2023generative}. In embodied and open-ended environments, skill libraries and reusable policies have also been shown to support long-horizon planning and continual exploration \citep{ahn2022saycan,liang2022codeaspolicies,wang2023voyager}. However, most existing skill-based agents either rely on manually specified skills, skills learned through environment interaction, or executable code generated during task solving. They do not directly address a complementary and practical source of capabilities: the large amount of latent procedural knowledge already embedded in heterogeneous external knowledge bases.

In this paper, we propose \textsc{Anything2Skill}, a taxonomy-guided framework that compiles heterogeneous external knowledge into reusable, retrievable, and executable skills for LLM agents. Instead of using external knowledge only as task-time context, \textsc{Anything2Skill} transforms documents, manuals, dialogues, logs, and trajectories into structured skill contracts. Each skill contract specifies the skill name, description, normalized asset type, taxonomy node, invocation conditions, contraindications, action moves, workflow steps, constraints, output specifications, supporting evidence, and confidence scores. These skills are stored in a persistent SkillBank, which serves as procedural memory for downstream agents.

The key idea behind \textsc{Anything2Skill} is to bridge retrieval-augmented knowledge access and skill-based capability reuse. Given a heterogeneous knowledge corpus, our framework first decomposes each record into evidence windows and performs plan-and-expand skill extraction under a skill-tree prior. The skill-tree prior constrains extraction toward procedural, evidence-grounded capabilities rather than free-form summaries. Extracted candidates are then normalized into stable asset types and converted into structured skill drafts. To make the resulting skills deployable, \textsc{Anything2Skill} further manages them in a SkillBank through taxonomy-aware compilation, registry-level reconciliation, lifecycle tracking, versioned updates, and visible skill-tree projection. At inference time, an agent retrieves both task-specific evidence from the original knowledge base and relevant procedural skills from the SkillBank. In this way, RAG provides declarative grounding, while the SkillBank provides reusable procedural guidance.

We evaluate \textsc{Anything2Skill} on agent tasks involving qsv and GitHub-CLI. Experimental results show that combining \textsc{Anything2Skill} with RAG achieves success rates of 98.85\% and 94.10\%, respectively, substantially outperforming RAG-only agents. These results indicate that many useful agent capabilities can be recovered from existing knowledge sources when implicit procedures are compiled into explicit, reusable skills. More broadly, our findings suggest that retrieval-augmented agents should not only retrieve external knowledge, but also organize and reuse the procedural capabilities latent within that knowledge.

Our contributions are summarized as follows:
\begin{itemize}[leftmargin=*, align=left]
    \item We propose \textsc{Anything2Skill}, a taxonomy-guided framework that extends retrieval-augmented agents from external knowledge access to reusable capability acquisition by compiling heterogeneous knowledge records into structured, evidence-grounded skills.

    \item We implement a complete skill compilation and management pipeline, including evidence-window construction, plan-and-expand skill extraction under a skill-tree prior, structured skill-contract generation, taxonomy-aware consolidation, registry-level reconciliation, lifecycle tracking, versioned updates, and visible SkillBank projection.

    \item We empirically demonstrate that augmenting RAG agents with the compiled SkillBank substantially improves task success on qsv and GitHub-CLI benchmarks, achieving 98.85\% and 94.10\% success rates, respectively, and consistently outperforming RAG-only agents.
\end{itemize}

\section{Related Work}
\label{sec:related_work}

\subsection{Retrieval-Augmented Generation}
Retrieval-augmented generation (RAG) has become a standard paradigm for equipping language models with external, updateable, and interpretable knowledge. Early open-domain question answering systems combined retrieval with neural reading models, retrieving evidence from large corpora before extracting or generating answers \citep{chen2017drqa,lee2019latent,seo2019real}. Dense Passage Retrieval (DPR) further demonstrated that learned dense retrievers can substantially improve open-domain evidence retrieval over sparse lexical baselines \citep{karpukhin2020dpr}, while REALM introduced retrieval into language model pre-training by learning a latent retriever over a large textual corpus \citep{guu2020realm}. RAG formalized the integration of a parametric sequence-to-sequence generator with a non-parametric memory accessed through dense retrieval, showing strong performance on knowledge-intensive NLP tasks \citep{lewis2020rag}. Subsequent work improved the reader-generator component, such as Fusion-in-Decoder, which encodes multiple retrieved passages independently and fuses them during decoding \citep{izacard2021fid}.

A second line of work studies retrieval as a scalable memory mechanism for language models. RETRO conditions autoregressive generation on chunks retrieved from a large-scale database, showing that retrieval can reduce the need to store all knowledge in model parameters \citep{borgeaud2022retro}. Atlas demonstrates that retrieval-augmented language models can support strong few-shot learning on knowledge-intensive tasks with an updateable document index \citep{izacard2023atlas}. REPLUG explores retrieval augmentation for black-box language models by prepending retrieved documents to the input and using the language model to supervise the retriever \citep{shi2024replug}. Other work investigates when parametric knowledge is sufficient and when non-parametric retrieval is necessary \citep{mallen2023when}, as well as benchmarks for evaluating knowledge-intensive language tasks \citep{petroni2021kilt}. Retrieval quality itself has also been improved through sparse-dense retrieval, late interaction, and unsupervised dense retrieval methods, including BM25-style probabilistic retrieval \citep{robertson2009probabilistic}, ColBERT \citep{khattab2020colbert}, Contriever \citep{izacard2022contriever}, and HyDE \citep{gao2023hyde}.

Recent RAG research moves beyond a single retrieve-then-generate pipeline. WebGPT retrieves and browses web evidence to support long-form question answering with human feedback \citep{nakano2021webgpt}. IRCoT interleaves retrieval with chain-of-thought reasoning, allowing intermediate reasoning steps to guide subsequent retrieval for multi-hop questions \citep{trivedi2023ircot}. Demonstrate-Search-Predict formulates retrieval-augmented in-context learning as a compositional pipeline over language-model and retrieval-model calls \citep{khattab2022dsp}. FLARE actively decides when and what to retrieve during generation, especially for long-form outputs \citep{jiang2023flare}. Self-RAG further trains models to decide when to retrieve, how to use retrieved passages, and how to critique their own generations through reflection tokens \citep{asai2024selfrag}. These methods improve the timing, grounding, and controllability of retrieval.

Despite these advances, most RAG systems still treat external knowledge as declarative evidence: passages, documents, examples, web pages, or API descriptions are retrieved and then interpreted at inference time. This design is effective for factual grounding, but it leaves the agent to repeatedly reconstruct procedures from fragmented context. In contrast, \textsc{Anything2Skill} focuses on the procedural knowledge implicitly embedded in heterogeneous knowledge bases. Rather than only retrieving passages, it compiles documents, manuals, logs, conversations, and trajectories into structured, evidence-grounded skills that can be stored, managed, retrieved, and reused. In this sense, our work complements RAG: retrieval provides task-time declarative evidence, while skill compilation provides reusable procedural memory.

\subsection{Agent Skills}
Large language models are increasingly used as agents that reason, call tools, interact with environments, and execute multi-step tasks. ReAct shows that interleaving reasoning traces with actions improves both question answering and interactive decision making \citep{yao2023react}. Toolformer trains language models to decide when and how to call external tools using self-supervised API-use data \citep{schick2023toolformer}. MRKL systems propose a modular neuro-symbolic architecture in which language models route problems to external tools, knowledge sources, or symbolic modules \citep{karpas2022mrkl}. HuggingGPT uses an LLM as a controller to plan tasks, select models from Hugging Face, execute subtasks, and summarize results \citep{shen2023hugginggpt}. Gorilla improves API calling by combining fine-tuning with retrieval over changing API documentation \citep{patil2024gorilla}. ToolLLM and ToolBench scale tool-use training and evaluation to thousands of real-world APIs \citep{qin2024toolllm}, while API-Bank provides a benchmark for tool-augmented LLMs \citep{li2023api}. Program-aided language models and program-of-thought prompting similarly externalize computation into executable programs \citep{gao2023pal,chen2022program}.

Another line of work studies agents that improve through feedback, memory, planning, and interaction. Reflexion enables language agents to use verbal feedback and maintain reflective memory without updating model weights \citep{shinn2023reflexion}. Generative Agents simulate human-like behavior by storing and retrieving memory records for planning and reflection \citep{park2023generative}. CAMEL studies communicative multi-agent role-playing as a way to elicit cooperative problem solving among language agents \citep{li2023camel}. Tree of Thoughts extends prompting with search over intermediate reasoning states \citep{yao2023tree}. These works demonstrate that agents benefit from externalized reasoning traces, memories, and feedback. However, their memories are often episodic, task-specific, or unstructured, whereas our work aims to compile reusable procedural skills from heterogeneous prior knowledge.

Skill-based agents have also been widely explored in robotics, embodied AI, and reinforcement learning. SayCan grounds language-model planning in pretrained robotic skills and affordance scores, enabling feasible long-horizon robotic behavior \citep{ahn2022saycan}. Inner Monologue uses natural-language feedback from the environment to support embodied reasoning and planning \citep{huang2022inner}. Code as Policies generates executable robot policy code from language instructions \citep{liang2022codeaspolicies}. ProgPrompt, LM-Nav, PaLM-E, RT-2, and language-model zero-shot planners further show that language models can compose actions, policies, and embodied knowledge for robotics and navigation \citep{singh2023progprompt,shah2023lmnav,driess2023palm-e,brohan2023rt2,huang2022language}. In open-ended environments, Voyager maintains an ever-growing library of executable code skills and reuses them for lifelong exploration in Minecraft \citep{wang2023voyager}. More broadly, reinforcement learning has studied skill discovery, option learning, and skill priors as mechanisms for temporal abstraction and generalization \citep{eysenbach2019diayn,pertsch2021spirl,andreas2017modular}, while program synthesis work such as DreamCoder learns reusable libraries that accelerate future problem solving \citep{ellis2021dreamcoder}. Recent work also explores inducing and planning with latent language skills \citep{sharma2022skill}.

These studies show that reusable skills are crucial for long-horizon agency, tool use, embodied control, and continual improvement. However, most existing agent-skill systems either assume that skills are manually specified, learned through environment interaction, distilled from execution attempts, or stored as code generated during task solving. \textsc{Anything2Skill} addresses a complementary setting: many useful procedures already exist implicitly in documents, manuals, dialogues, logs, and trajectories, but they are not directly available as reusable agent skills. Our framework extracts such latent procedural knowledge, grounds it in source evidence, normalizes it with a skill taxonomy, and manages it in a persistent SkillBank through compilation, reconciliation, versioning, and hierarchy projection. This allows agents to retrieve not only task-relevant content, but also reusable procedural capabilities compiled from external knowledge.

\section{Method}
\label{sec:method}

\subsection{Overview}
\label{sec:method_overview}

\begin{figure}
    \centering
    \includegraphics[width=0.95\linewidth]{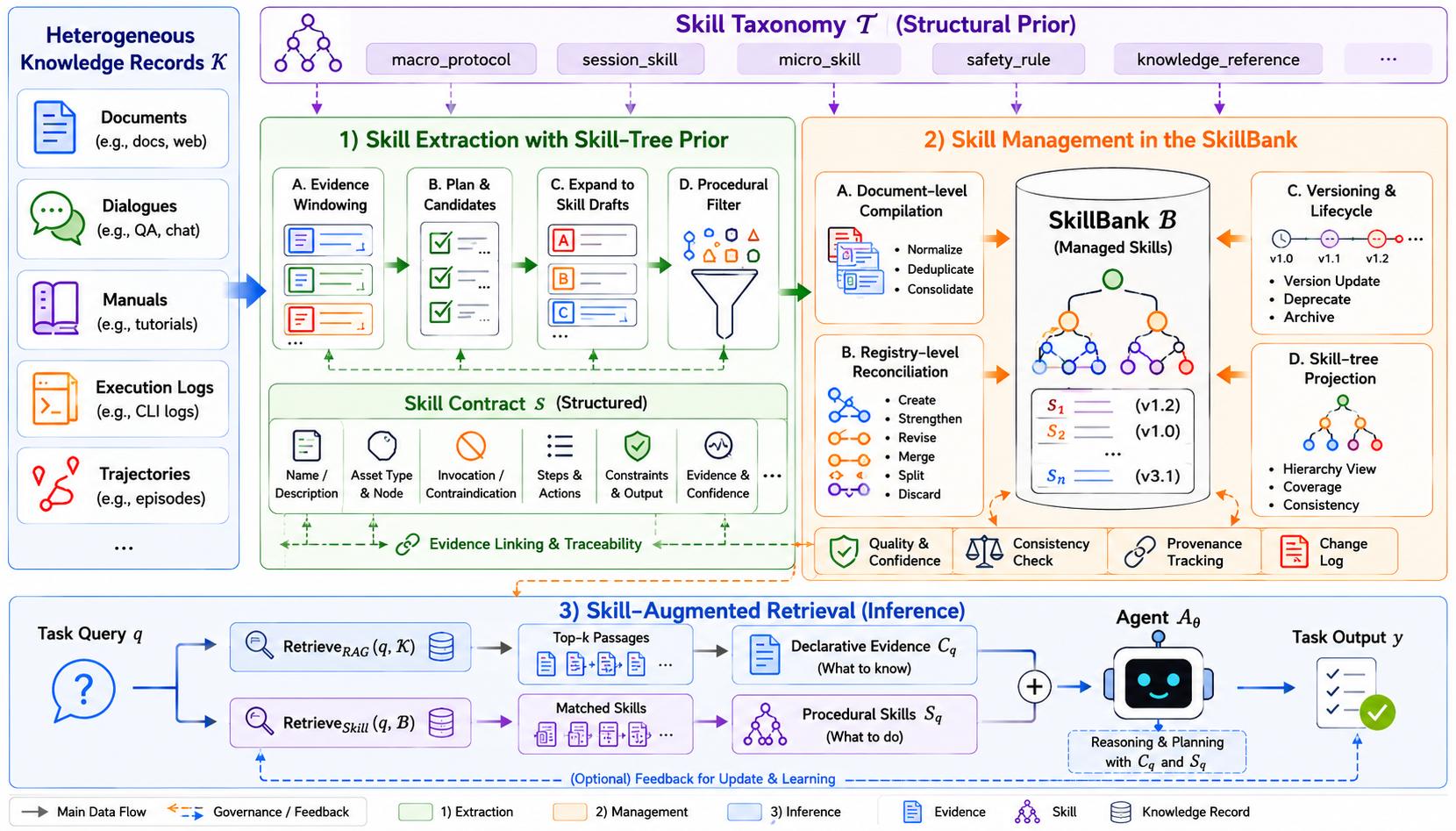}
    \caption{Overview of \textsc{Anything2Skill}, which compiles heterogeneous knowledge records into reusable procedural skills. The framework uses a skill taxonomy as a structural prior for skill extraction, manages extracted skills through compilation, reconciliation, and versioned tree projection, and supports inference by retrieving both declarative evidence and procedural skills for agent execution.}
    \label{fig:framework}
\end{figure}

Retrieval-augmented generation enables agents to access external knowledge at inference time, but it primarily exposes agents to fragmented declarative evidence, leaving them to repeatedly reconstruct task procedures from passages, examples, manuals, logs, or trajectories. \textsc{Anything2Skill} addresses this limitation by compiling heterogeneous external knowledge into reusable, retrievable, and executable skills (Figure \ref{fig:framework}). Given a corpus of knowledge records $\mathcal{K}=\{D_i\}_{i=1}^{N}$, where each $D_i$ may be a document, dialogue, manual, execution log, or trajectory, our goal is to discover latent procedural knowledge that specifies how tasks should be performed, when tools or actions should be invoked, what constraints should be followed, and what outputs should be produced. The compiled skills are stored in a persistent SkillBank $\mathcal{B}$ and can be retrieved at inference time to guide agent execution, thereby complementing RAG with procedural capabilities distilled from external knowledge bases.

Formally, \textsc{Anything2Skill} performs a taxonomy-guided compilation process:
\begin{equation}
\mathcal{B}
=
\textsc{Anything2Skill}(\mathcal{K},\mathcal{T}),
\end{equation}
where $\mathcal{T}$ is a skill taxonomy that provides a structural prior over skill asset types, hierarchy levels, and parent-child relations. The method consists of two main components. First, \textit{skill extraction with a skill-tree prior} identifies evidence-grounded procedural knowledge from heterogeneous records and converts it into structured skill drafts. Second, \textit{skill management in the SkillBank} consolidates, reconciles, versions, and organizes extracted skills into a coherent repository for later retrieval and execution.

Each skill $s$ is represented as a structured contract:
\begin{equation}
\label{eq:skill_contract}
s =
\langle
n, d, a, v, g, \tau,
\sigma^{+}, \sigma^{-},
\mathbf{m}, \mathbf{p}, \mathbf{c}, \mathbf{o},
\mathcal{E}, \rho
\rangle,
\end{equation}
where $n$ and $d$ are the skill name and description; $a$ is the normalized asset type; $v$ is the taxonomy node; $g$ is the granularity level; $\tau$ stores contextual metadata such as domain, family, task type, method type, and stage; $\sigma^{+}$ and $\sigma^{-}$ denote invocation and contraindication conditions; $\mathbf{m}$ denotes action or intervention moves; $\mathbf{p}$ denotes workflow steps; $\mathbf{c}$ denotes constraints and cautions; $\mathbf{o}$ denotes the output contract; $\mathcal{E}$ is the supporting evidence set; and $\rho$ is the confidence score.

\subsection{Skill Extraction with Skill-Tree Prior}
\label{sec:skill_extraction}

\paragraph{Skill taxonomy.}
The skill taxonomy $\mathcal{T}$ defines the structural prior used to guide extraction and later management. We define
\begin{equation}
\mathcal{T}
=
(\mathcal{A},\mathcal{V},\mathcal{R},\pi,\ell,\mathcal{F}),
\end{equation}
where $\mathcal{A}$ is the set of stable internal asset types, $\mathcal{V}$ is the set of taxonomy nodes, $\mathcal{R}\subseteq\mathcal{V}\times\mathcal{V}$ defines parent-child relations, $\pi:\mathcal{V}\rightarrow\mathcal{A}$ maps each taxonomy node to an asset type, $\ell:\mathcal{V}\rightarrow\mathbb{N}$ assigns a hierarchy level, and $\mathcal{F}$ stores domain- or family-specific buckets.

The internal asset type space is intentionally compact:
\begin{equation}
\mathcal{A}
=
\left\{
\texttt{macro\_protocol},
\texttt{session\_skill},
\texttt{micro\_skill},
\texttt{safety\_rule},
\texttt{knowledge\_reference}
\right\}.
\end{equation}
Domain-specific aliases are normalized into this stable type space through a taxonomy-specific normalizer:
\begin{equation}
a = N_{\mathcal{T}}(\tilde{a}),
\end{equation}
where $\tilde{a}$ is the raw asset label produced by the extractor. If $N_{\mathcal{T}}(\tilde{a})=\bot$, the candidate is rejected.

Given a candidate skill $\tilde{s}$ and its normalized asset type $a$, its taxonomy node is resolved as
\begin{equation}
v
=
\arg\max_{v'\in\mathcal{V}:\pi(v')=a}
\operatorname{match}(\tilde{s},v'),
\end{equation}
where $\operatorname{match}(\cdot)$ measures compatibility between the candidate and a taxonomy node using node labels, aliases, hierarchy constraints, and family metadata. This taxonomy prior constrains the extractor to produce procedural skills rather than free-form summaries, and prevents incompatible skill types from being merged in later management.

\paragraph{Evidence windows.}
Each input record $D_i$ is first normalized and decomposed into strict evidence windows:
\begin{equation}
W_i
=
\{w_{i1},w_{i2},\ldots,w_{im_i}\}.
\end{equation}
Each window is represented as
\begin{equation}
w_{ij}
=
\langle
x_{ij}, h_{ij}, r_{ij}, b_{ij}, m_{ij}
\rangle,
\end{equation}
where $x_{ij}$ is the text span, $h_{ij}$ is the local heading context, $r_{ij}$ denotes the original paragraph or character span, $b_{ij}$ records boundary and anchor information, and $m_{ij}$ stores metadata such as source path, heading path, sibling headings, and subsection information. This design ensures that every extracted skill can be traced back to explicit source evidence.

\paragraph{Plan-and-expand extraction.}
Directly asking a model to extract all skills from a long document often produces redundant, underspecified, or summary-like outputs. \textsc{Anything2Skill} therefore uses a two-stage plan-and-expand extraction strategy. In the planning stage, an LLM-based planner identifies distinct skill candidates from each evidence window:
\begin{equation}
\mathcal{P}_{ij}
=
P_{\theta}(w_{ij},\mathcal{T},K),
\end{equation}
where $K$ is the maximum number of candidates per window. Each planned candidate $p_{ijk}\in\mathcal{P}_{ij}$ contains a lightweight objective, tentative asset type, taxonomy node, invocation cues, contraindications, expected artifact, evidence hints, and confidence.

In the expansion stage, each candidate is independently expanded into a structured skill draft:
\begin{equation}
\tilde{s}_{ijk}
=
E_{\theta}(w_{ij},p_{ijk},\mathcal{T}).
\end{equation}
The expansion is conservative: if the window does not directly support the planned procedure, the extractor returns a null draft rather than hallucinating unsupported steps.

To distinguish procedural skills from ordinary factual descriptions, we apply a procedural validity filter:
\begin{equation}
\label{eq:procedural_filter}
I_{\mathrm{proc}}(\tilde{s})
=
\mathbb{I}
\left[
|\mathbf{p}|+|\mathbf{m}|+|\mathbf{c}|>0
\right]
\cdot
\mathbb{I}
\left[
N_{\mathcal{T}}(\tilde{a})\neq\bot
\right].
\end{equation}
A candidate is retained only if it contains at least one procedural component, such as workflow steps, action moves, constraints, or cautions, and can be mapped into the taxonomy.

For each retained draft, \textsc{Anything2Skill} creates a support record:
\begin{equation}
e_{ijk}
=
\langle
\operatorname{id}(D_i),
r_{ij},
\operatorname{excerpt}(w_{ij}),
q_{ijk},
\rho_{ijk}
\rangle,
\end{equation}
where $q_{ijk}$ denotes the support relation type, such as support, constraint, conflict, or case variant. The extracted draft is represented as
\begin{equation}
d_{ijk}
=
\langle
\tilde{s}_{ijk},
e_{ijk},
\operatorname{meta}(w_{ij},D_i,\mathcal{T})
\rangle.
\end{equation}
The extraction output for document $D_i$ is
\begin{equation}
\mathcal{D}_{i}
=
\{d_{ijk}\mid I_{\mathrm{proc}}(\tilde{s}_{ijk})=1\}.
\end{equation}

\subsection{Skill Management in the SkillBank}
\label{sec:skill_management}

Skill extraction produces local drafts, while a deployable agent requires a coherent, non-redundant, and versioned SkillBank. \textsc{Anything2Skill} therefore performs skill management in three steps: document-level skill compilation, registry-level reconciliation, and visible skill-tree projection.

\paragraph{Document-level skill compilation.}
Window-level drafts may describe overlapping parts of the same capability. To produce canonical document-level skills, \textsc{Anything2Skill} groups drafts using a taxonomy-aware identity key.
For a draft record $d$, let $s_d=\operatorname{skill}(d)$ denote its skill payload and $D_d=\operatorname{src}(d)$ denote its source record. Field accessors such as $a(\cdot)$, $v(\cdot)$, and $\tau_{\mathrm{domain}}(\cdot)$ refer to fields of a skill contract.
\begin{equation}
\label{eq:compile_key}
\begin{aligned}
\gamma(d)
=
\bigl(&
\operatorname{id}(D_d),
a(s_d),
g(s_d),
v(s_d),
\\
&
\operatorname{norm}(n(s_d)),
\operatorname{norm}(\mathbf{o}(s_d)),
\tau_{\mathrm{domain}}(s_d),
\tau_{\mathrm{family}}(s_d),
\\
&
\tau_{\mathrm{task}}(s_d),
\tau_{\mathrm{method}}(s_d),
\tau_{\mathrm{stage}}(s_d)
\bigr).
\end{aligned}
\end{equation}
Drafts sharing the same key are treated as document-local descriptions of the same capability. For each group
\begin{equation}
G_c=\{d\mid \gamma(d)=c\},
\end{equation}
a compiler produces a canonical skill specification:
\begin{equation}
\label{eq:compile_group}
s_c
=
C_{\theta}(G_c,\mathcal{E}_{G_c}),
\end{equation}
where
\begin{equation}
\mathcal{E}_{G_c}
=
\bigcup_{d\in G_c}\mathcal{E}(d).
\end{equation}
The compiler is constrained to preserve the group-level asset type, granularity, and taxonomy node:
\begin{equation}
\label{eq:compile_constraint}
\forall d\in G_c,\quad
a(s_c)=a(s_d),\quad
g(s_c)=g(s_d),\quad
v(s_c)=v(s_d).
\end{equation}
It is also required to cite only evidence contained in $\mathcal{E}_{G_c}$:
\begin{equation}
\mathcal{E}(s_c)
\subseteq
\mathcal{E}_{G_c}.
\end{equation}
Conceptually, compilation solves a constrained canonicalization objective:
\begin{equation}
\label{eq:canonical_objective}
s_c^{*}
=
\arg\max_{s}
\left[
\sum_{d\in G_c}\operatorname{cover}(s,d)
-
\lambda\operatorname{unsupported}(s,\mathcal{E}_{G_c})
-
\mu\operatorname{redundancy}(s)
\right],
\end{equation}
subject to Eq.~\ref{eq:compile_constraint}. This objective encourages the canonical skill to preserve supported procedural content while avoiding unsupported or redundant claims.

\paragraph{Registry-level reconciliation.}
After compilation, each incoming skill is reconciled with the persistent SkillBank. For an incoming skill $s$, let $\mathcal{S}_{\mathrm{reg}}$ denote the canonical skills currently stored in the registry. \textsc{Anything2Skill} retrieves related existing skills:
\begin{equation}
H_k(s)
=
\operatorname{TopK}_{t\in\mathcal{S}_{\mathrm{reg}}}R(s,t),
\end{equation}
where $R(s,t)$ is a hybrid retrieval score combining dense similarity, sparse lexical matching, and taxonomy-aware compatibility:
\begin{equation}
\label{eq:hybrid_score}
\begin{aligned}
R(s,t)
={}&
(1-\alpha)\cos(\phi(s),\phi(t))
+
\alpha\operatorname{BM25}(x_s,x_t)
\\
&
+
\beta_1\mathbb{I}[a(s)=a(t)\land g(s)=g(t)]
+
\beta_2\mathbb{I}[v(s)=v(t)]
\\
&
+
\beta_3\mathbb{I}[\ell(v(t))<\ell(v(s))].
\end{aligned}
\end{equation}
Here $\phi(\cdot)$ is the dense embedding of a metadata-rich skill text, and $x_s,x_t$ are sparse retrieval texts built from skill names, descriptions, objectives, triggers, workflows, constraints, outputs, and taxonomy metadata.

The BM25 component is defined as
\begin{equation}
\label{eq:bm25}
\operatorname{BM25}(q,d)
=
\frac{1}{Z}
\sum_{u\in q}
\operatorname{idf}(u)
\cdot
\frac{
f(u,d)(k_1+1)
}{
f(u,d)+k_1(1-b+b|d|/\overline{|d|})
},
\end{equation}
where
\begin{equation}
\operatorname{idf}(u)
=
\log
\left(
1+
\frac{N-n_u+0.5}{n_u+0.5}
\right).
\end{equation}
Here, the second BM25 argument $d$ denotes a sparse retrieval text; the draft record variable in document-level compilation has local scope.

Given the retrieved neighbors $H_k(s)$, peer incoming skills $\mathcal{S}_{\mathrm{peer}}(s)$, and support evidence $\mathcal{E}(s)$, a management model predicts a lifecycle action:
\begin{equation}
\label{eq:management_action}
\eta
=
M_{\theta}
\left(
s,
H_k(s),
\mathcal{S}_{\mathrm{peer}}(s),
\mathcal{E}(s)
\right),
\end{equation}
where
\begin{equation}
\eta
\in
\left\{
\textsc{Create},
\textsc{Strengthen},
\textsc{Revise},
\textsc{Merge},
\textsc{Split},
\textsc{Unchanged},
\textsc{Discard}
\right\}.
\end{equation}
The action space is taxonomy-constrained. For a selected existing skill $t\in H_k(s)$, update-like actions require taxonomy compatibility:
\begin{equation}
\label{eq:merge_constraint}
\begin{aligned}
\eta
&\in
\left\{
\textsc{Strengthen},
\textsc{Revise},
\textsc{Merge},
\textsc{Unchanged}
\right\}
\\
&\Rightarrow
a(s)=a(t)
\land
g(s)=g(t)
\\
&\qquad\land
(v(s)=v(t)\lor v(s)=\emptyset\lor v(t)=\emptyset).
\end{aligned}
\end{equation}
This prevents skills from being merged across incompatible asset layers, such as merging a safety rule into an executable micro-skill.

The registry applies the predicted action as follows. \textsc{Create} inserts a new skill. \textsc{Strengthen} preserves the existing skill identity while adding new evidence and supported fields. \textsc{Revise} updates the skill body when new evidence changes the procedure. \textsc{Merge} combines compatible duplicate skills and deprecates secondary records. \textsc{Split} creates or links a narrower child skill. \textsc{Unchanged} records provenance without changing the skill body. \textsc{Discard} removes unsupported or non-procedural candidates.

Each skill maintains a lifecycle state:
\begin{equation}
z(s)
\in
\left\{
\texttt{candidate},
\texttt{draft},
\texttt{evaluating},
\texttt{active},
\texttt{watchlist},
\texttt{deprecated},
\texttt{retired}
\right\}.
\end{equation}
Versioned updates are recorded as
\begin{equation}
\label{eq:version_update}
\operatorname{ver}'(s)
=
U
\left(
\operatorname{ver}(s),
\eta,
\Delta\mathcal{E},
\Delta s
\right),
\end{equation}
where $\Delta\mathcal{E}$ denotes newly attached evidence and $\Delta s$ denotes the content change induced by the lifecycle action.

\paragraph{Visible skill-tree projection.}
The registry is the source of truth, while the visible skill tree is a rebuildable projection for navigation, inspection, and retrieval. After registry reconciliation, active skills are organized into a domain/family/level hierarchy according to $\mathcal{T}$. Let $\mathcal{S}_{\mathrm{act}}\subseteq\mathcal{S}_{\mathrm{reg}}$ denote the active canonical skills in the registry. For a skill $s$, we define its taxonomy level as
\begin{equation}
\operatorname{level}(s)=\ell(v(s)).
\end{equation}
For any skill $s$ with $\operatorname{level}(s)>1$, candidate parents are selected from active canonical skills whose taxonomy nodes are valid parents of $v(s)$:
\begin{equation}
\mathcal{P}_{\mathrm{parent}}(s)
=
\{t\in\mathcal{S}_{\mathrm{act}}:
\operatorname{level}(t)=\operatorname{level}(s)-1
\land
(v(t),v(s))\in\mathcal{R}
\}.
\end{equation}
The parent is selected as
\begin{equation}
\operatorname{parent}(s)
=
\arg\max_{t\in\mathcal{P}_{\mathrm{parent}}(s)}
R(s,t),
\end{equation}
subject to a confidence threshold. If no compatible parent is found, the skill remains available in the registry but is left unlinked in the visible tree.

To reduce navigational redundancy, the system may further perform same-level consolidation over projected sibling skills. Two skills are considered consolidation candidates when they occupy the same metadata and taxonomy slot:
\begin{equation}
\label{eq:same_level_key}
\kappa(s)
=
\left(
a(s),
g(s),
v(s),
\operatorname{level}(s),
\tau_{\mathrm{domain}}(s),
\tau_{\mathrm{family}}(s),
\tau_{\mathrm{task}}(s),
\tau_{\mathrm{method}}(s),
\tau_{\mathrm{stage}}(s),
\operatorname{norm}(n(s))
\right).
\end{equation}
For each duplicate group
\begin{equation}
G_{\kappa}
=
\{s\in\mathcal{S}_{\mathrm{act}}:\kappa(s)=\kappa\},
\end{equation}
the primary skill is selected as
\begin{equation}
\label{eq:primary_skill}
s^{*}
=
\arg\max_{s\in G_{\kappa}}Q(s),
\end{equation}
where $Q(s)$ ranks skills by lifecycle status, provenance coverage, content completeness, version maturity, and persistence in the registry. Non-conflicting fields are merged into the primary skill:
\begin{equation}
\label{eq:dedup_union}
F(s^{*})
=
\operatorname{DedupUnion}_{s\in G_{\kappa}}F(s),
\end{equation}
where $F$ ranges over evidence records, invocation signals, contraindications, workflow steps, constraints, output contracts, triggers, tags, examples, and child links. Secondary records are deprecated, and visible-tree links are rewritten through
\begin{equation}
\label{eq:rewrite_links}
\psi(s)
=
\begin{cases}
s^{*}, & \text{if } s\in G_{\kappa}\setminus\{s^{*}\},\\
s, & \text{otherwise}.
\end{cases}
\end{equation}
For any skill-level hierarchy edge $(s_p,s_c)$, the updated edge becomes
\begin{equation}
(s_p,s_c)\mapsto(\psi(s_p),\psi(s_c)).
\end{equation}

The final SkillBank is represented as
\begin{equation}
\mathcal{B}
=
\langle
\mathcal{S},
\mathcal{E}_{\mathrm{store}},
\mathcal{H},
\mathcal{V}_{\mathrm{hist}},
\mathcal{L}
\rangle,
\end{equation}
where $\mathcal{S}$ is the set of canonical skills, $\mathcal{E}_{\mathrm{store}}$ is the evidence store, $\mathcal{H}$ is the visible hierarchy projection, $\mathcal{V}_{\mathrm{hist}}$ stores version history, and $\mathcal{L}$ stores lifecycle and provenance logs.

\subsection{Skill-Augmented Retrieval}
\label{sec:skill_rag}

At inference time, the constructed SkillBank can be combined with a standard RAG pipeline. Given a task query $q$, the agent retrieves both task-time evidence passages and relevant procedural skills:
\begin{equation}
\mathcal{C}_{q}
=
\operatorname{Retrieve}_{\mathrm{RAG}}(q,\mathcal{K}),
\qquad
\mathcal{S}_{q}
=
\operatorname{Retrieve}_{\mathrm{Skill}}(q,\mathcal{B}).
\end{equation}
The downstream agent then solves the task conditioned on both sources:
\begin{equation}
y
=
A_{\theta}
\left(
q,
\mathcal{C}_{q},
\mathcal{S}_{q}
\right).
\end{equation}
Here, RAG provides task-specific declarative evidence, while \textsc{Anything2Skill} provides reusable procedural memory distilled from previous knowledge records.

\section{Experiments}
\label{sec:experiments}

\subsection{Experimental Setup}
\label{sec:experimental_setup}

We evaluate \textsc{Anything2Skill} by asking two research questions. First, can compiled skill contracts improve agent performance even without task-time document retrieval? Second, can compiled skills complement retrieval-augmented generation (RAG) when both procedural skills and source passages are available \citep{lewis2020rag}? To answer these questions, we conduct experiments on two command-line agent benchmarks and report task success rate as the main evaluation metric.

\paragraph{Benchmarks.}
We use two command-line benchmarks that require agents to understand tool documentation, compose commands, and execute multi-step operations. The qsv benchmark focuses on CSV command-line operations grounded in the official qsv documentation \citep{qsv_docs}. Tasks in this benchmark require the agent to select appropriate commands, compose options, and perform tabular-data transformations. The GitHub-CLI benchmark is constructed from GitHub CLI repository operation guides and \texttt{gh} command documentation \citep{github_cli_manual}. It covers 110 source documents and evaluates repository-management operations such as issue handling, pull-request workflows, release management, and repository inspection. From this corpus, \textsc{Anything2Skill} compiles 179 GitHub-CLI skills, including 1 global skill, 6 top-level skills, 62 second-level skills, and 110 micro-skills.

\paragraph{Compared configurations.}
We compare four context configurations. \textbf{Base Agent} denotes the setting where the agent receives no additional external context. \textbf{\textsc{Anything2Skill}} provides the agent with compiled skills retrieved from the SkillBank, but does not retrieve task-time source passages. \textbf{RAG} retrieves the top-3 source entries for each task from the original knowledge base. \textbf{\textsc{Anything2Skill} + RAG} combines both sources, where RAG provides task-specific declarative evidence and the SkillBank provides reusable procedural guidance. All configurations use the same base model, \texttt{gpt5.4-mini}, and are evaluated under the same task protocol.

\paragraph{Metric.}
We report benchmark success rate, defined as the percentage of tasks for which the agent successfully completes the requested command-line operation. This metric directly reflects whether the agent can translate a natural-language task into a valid and executable command-line solution.

\begin{table}[t]
  \centering
  \small
  \caption{Success rates on two command-line agent benchmarks. \textsc{Anything2Skill} improves agent performance by providing reusable procedural skills, and achieves the best results when combined with RAG.}
  \label{tab:cli-benchmark-results}
  \begin{tabular}{lcc}
    \toprule
    Method & qsv & GitHub-CLI \\
    \midrule
    Base Agent & 81.60\% & 64.70\% \\
    RAG & 95.41\% & 76.50\% \\
    \midrule
    \textsc{Anything2Skill} & 91.95\% & 82.30\% \\
    \textsc{Anything2Skill} + RAG & \textbf{98.85\%} & \textbf{94.10\%} \\
    \bottomrule
  \end{tabular}
\end{table}

\subsection{Experimental Analysis}
\label{sec:experimental_analysis}

\paragraph{Compiled skills improve agents without task-time retrieval.}
As shown in Table~\ref{tab:cli-benchmark-results}, \textsc{Anything2Skill} substantially improves over the Base Agent on both benchmarks. On qsv, the success rate increases from 81.60\% to 91.95\%, yielding an absolute gain of 10.35 percentage points. On GitHub-CLI, the improvement is larger, from 64.70\% to 82.30\%, corresponding to an absolute gain of 17.60 percentage points. These results show that the compiled SkillBank is not merely an additional retrieval index, but provides reusable procedural knowledge that can guide the agent even when no task-time document passages are retrieved.

\paragraph{Skills and RAG provide complementary context.}
RAG alone achieves strong performance on qsv, improving the Base Agent from 81.60\% to 95.41\%. This suggests that qsv tasks often benefit from directly retrieving command documentation and option descriptions. However, adding \textsc{Anything2Skill} on top of RAG further improves the success rate to 98.85\%, indicating that procedural skill contracts provide useful guidance beyond raw source passages. The same complementarity is even more pronounced on GitHub-CLI: RAG alone reaches 76.50\%, while \textsc{Anything2Skill} + RAG achieves 94.10\%. This demonstrates that declarative evidence and procedural skills address different aspects of agent execution. RAG helps the agent access task-specific facts, while \textsc{Anything2Skill} helps the agent reuse structured procedures such as invocation conditions, action sequences, constraints, and output expectations.

\paragraph{Procedural skills are especially useful for complex CLI workflows.}
The gain from \textsc{Anything2Skill} is larger on GitHub-CLI than on qsv. One possible reason is that GitHub-CLI tasks involve more diverse repository-management workflows, where the agent must identify the appropriate command family, satisfy command-specific constraints, and compose multiple options under different scenarios. In such cases, simply retrieving documentation may expose relevant fragments, but the agent still needs to infer the correct procedure from them. By contrast, \textsc{Anything2Skill} compiles these latent procedures into structured skill contracts before inference, reducing the burden of task-time procedural reconstruction. This explains why \textsc{Anything2Skill} alone outperforms RAG alone on GitHub-CLI, and why their combination achieves the strongest result.

\paragraph{Overall effectiveness.}
Across both benchmarks, \textsc{Anything2Skill} + RAG consistently achieves the best performance, with 98.85\% success on qsv and 94.10\% success on GitHub-CLI. Compared with the Base Agent, the combined setting improves success rate by 17.25 percentage points on qsv and 29.40 percentage points on GitHub-CLI. These results support the central hypothesis of this paper: retrieval-augmented agents should not only retrieve external documents as declarative evidence, but also compile the procedural knowledge latent in those documents into reusable skills. The resulting SkillBank complements RAG and enables agents to move from knowledge access toward capability reuse.

\section{Conclusion}
\label{sec:conclusion}

In this paper, we introduced \textsc{Anything2Skill}, a taxonomy-guided framework that extends retrieval-augmented agents from external knowledge access to reusable capability acquisition. Unlike conventional RAG systems that primarily retrieve fragmented declarative evidence at inference time, \textsc{Anything2Skill} compiles heterogeneous knowledge records, including documents, manuals, dialogues, logs, and trajectories, into structured, evidence-grounded skill contracts. These skills are further consolidated, reconciled, versioned, and organized in a persistent SkillBank, enabling agents to retrieve procedural guidance alongside task-specific source passages. Experiments on qsv and GitHub-CLI demonstrate that compiled skills substantially improve agent success, and that combining \textsc{Anything2Skill} with RAG achieves the best performance across both benchmarks. These results suggest that the procedural knowledge latent in existing knowledge bases can be transformed into reusable agent capabilities, offering a promising direction for building retrieval-augmented agents that not only access knowledge, but also accumulate, manage, and reuse skills over time.

\bibliographystyle{unsrtnat}
\nocite{yang2026autoskill}
\bibliography{references}

\end{document}